%% file: root.tex
\documentclass[letterpaper, 10 pt, conference]{ieeeconf}  

\IEEEoverridecommandlockouts                              

\usepackage{cite}
\usepackage{amsmath,amssymb,amsfonts}
\usepackage{graphicx}
\usepackage{textcomp}
\usepackage{xcolor}
\usepackage{gensymb}
\usepackage{multirow}
\usepackage{subcaption}
\usepackage{commath} 
\usepackage{ushort} 
\usepackage{url} 

\usepackage{epsfig} 
\usepackage{balance} 

\usepackage{adjustbox}
\usepackage{booktabs, makecell, tabularx}
\setcellgapes{3.5pt}

\usepackage{algorithm} 
\usepackage[noend]{algpseudocode} 
\usepackage{footnote} 
\usepackage{float} 
\usepackage{kotex}
\usepackage{wrapfig}

\makeatletter
\let\NAT@parse\undefined
\makeatother
\usepackage[pagebackref,breaklinks,hidelinks]{hyperref}

\title{\LARGE \bf
Skill Q-Network: Learning Adaptive Skill Ensemble \\for Mapless Navigation in Unknown Environments
}

\author{Hyunki Seong$^{*1}$ and David Hyunchul Shim$^{1}$
\thanks{This work was supported by the Agency for Defense Development Grant funded by Korean Government(UE231060TD).}
\thanks{$^*$Corresponding author}
\thanks{$^{1}$Hyunki Seong and David Hyunchul Shim are with the School of Electrical Engineering, Korea Advanced Institute of Science and Technology, Daejeon, South Korea.
        {\tt\small \{hynkis, hcshim\}@kaist.ac.kr}
        }%
}

\begin{document}

\maketitle
\thispagestyle{empty}
\pagestyle{empty}


\begin{abstract}
\input{sections/0.abstract}
\end{abstract}

\input{sections/1.introduction.tex}
\input{sections/2.relatedworks.tex}
\input{sections/4.methodologies.tex}

\input{sections/5.experiment.tex}
\input{sections/6.conclusion.tex}

\addtolength{\textheight}{-12cm}   


\bibliographystyle{IEEEtran}
\bibliography{root}

\end{document}

%% file: sections/0.abstract.tex
This paper focuses on the acquisition of mapless navigation skills within unknown environments. We introduce the Skill Q-Network (SQN), a novel reinforcement learning method featuring an adaptive skill ensemble mechanism. Unlike existing methods, our model concurrently learns a high-level skill decision process alongside multiple low-level navigation skills, all without the need for prior knowledge. Leveraging a tailored reward function for mapless navigation, the SQN is capable of learning adaptive maneuvers that incorporate both exploration and goal-directed skills, enabling effective navigation in new environments. Our experiments demonstrate that our SQN can effectively navigate complex environments, exhibiting a 40\% higher performance compared to baseline models. Without explicit guidance, SQN discovers how to combine low-level skill policies, showcasing both goal-directed navigations to reach destinations and exploration maneuvers to escape from local minimum regions in challenging scenarios. Remarkably, our adaptive skill ensemble method enables zero-shot transfer to out-of-distribution domains, characterized by unseen observations from non-convex obstacles or uneven, subterranean-like environments. The project page is available at \url{https://sites.google.com/view/skill-q-net}.

%% file: sections/1.introduction.tex
\section{Introduction}
\label{sec:introduction}
Safe and efficient navigation in unknown environments remains a significant challenge in robotics. In scenarios where the robotic agent has access to global maps, finding the optimal path to the goal point is feasible. However, in unfamiliar environments, the agent must rely solely on partially observable, ego-centric information for navigation.

In navigation challenges lacking prior information, two key requirements emerge: 1) the application of adaptive navigation strategies suitable for the current situation, and 2) the adoption of a comprehensive approach to handle unseen, out-of-distribution (OOD) observations.
If the agent merely aims towards the goal, it could encounter dead-ends or repeatedly traverse previously visited areas, resulting in inefficient back-and-forth movements. Therefore, beyond a goal-directed strategy, the agent must also adaptively employ exploration skills to navigate out of local minima. Furthermore, in new environments, it is likely to encounter partially-known scenes and unfamiliar OOD terrains, necessitating a robust ability to manage novel situations.

Recent studies have introduced numerous experience-based reinforcement learning (RL) approaches~\cite{zhelo2018curiosity,lv2021deep} for navigating environments without maps. These approaches employ policy networks, trained through reward functions with minimal heuristics, to navigate to destinations using egocentric range information about the surrounding terrain and relative positioning information toward the goal. moreover, incorporating memory-based modules like LSTM or GRU enables the development of navigation policies that integrate the temporal characteristics of the ego robot's historical trajectory~\cite{dobrevski2021deep,wei2024memory}. However, learning various strategies with a single end-to-end network remains a challenge. Without pre-defining and individually training low-level skills~\cite{huang2023creating,lee2023adaptive}, it is still difficult to train a unified policy network that can employ a diverse set of navigation skills.

\begin{figure}[t]
\centering
\includegraphics[width=0.48\textwidth]{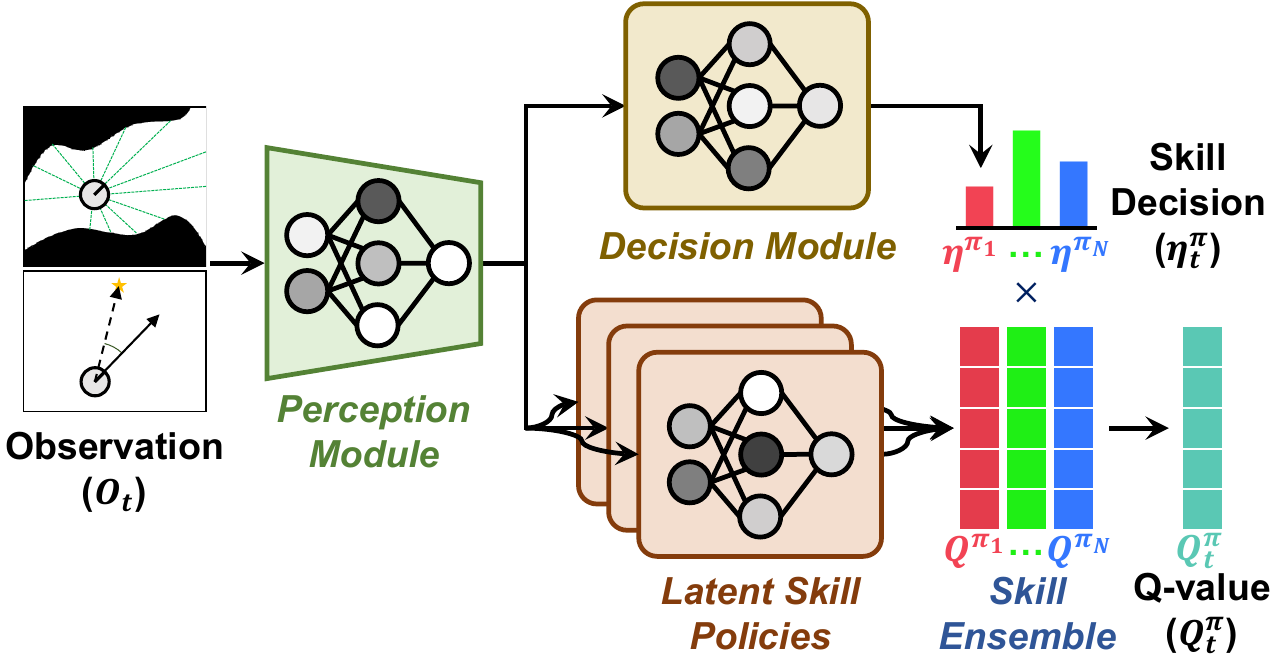}
\caption{
Overview of the Skill Q-Network.
}
\label{fig:intro}
\vspace{-1.5em}
\end{figure}

\begin{figure*}[t]
\centering
\includegraphics[width=0.95\textwidth]{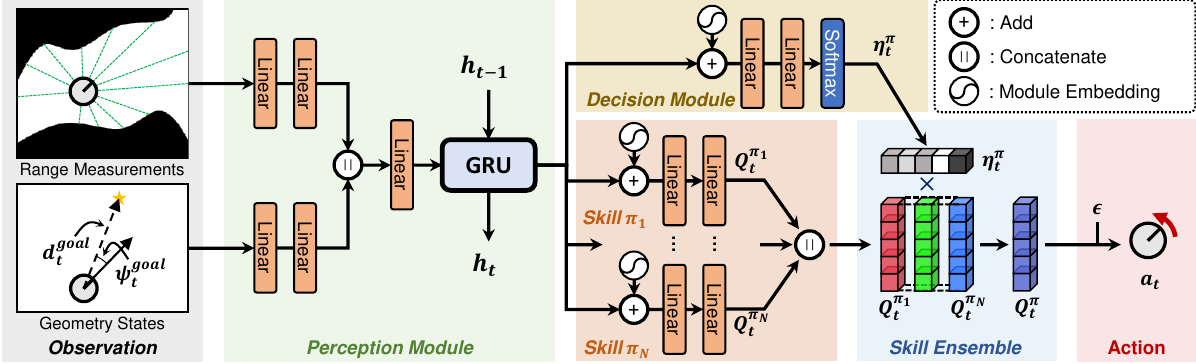}
\caption{
Detailed network architecture of the Skill Q-Network.
}
\label{fig:network}
\vspace{-1.5em}
\end{figure*}

In this paper, we introduce Skilled Q-Network (SQN), a novel deep Q-learning approach for end-to-end mapless navigation that incorporates an adaptive skill ensemble mechanism. Our network features multiple latent skill policies and a skill decision module, differentiated through module embedding processes. The decision module infers a skill decision, which evaluates importance scores for each skill, facilitating an internal high-level decision-making process within the end-to-end architecture. The skill decision then aggregates the Q-values from the latent skill policies into a single Q-value vector for action selection. This skill-ensembled Q-learning approach, with a latent skill decision mechanism, enables the network to learn adaptive navigation strategies without the need for prior skill-level knowledge.

To train navigation skills, we formulate a tailored reward function with terms that encourage exploration of unknown environments for discovering feasible goal achievement regions and exploitation of goal-reachable situations for successful arrival. By providing these reward signals that balance exploration and goal-directed features, we enable SQN to acquire a diverse set of navigation skills, adept at navigating complex, unknown environments.

In our extensive experiments, SQN consistently outperforms baseline models across various mapless environments, demonstrating a $40\%$ improvement in performance compared to conventional models. Additionally, we investigate the adaptive skill decision trajectories of our model, showcasing its ability to effectively combine latent navigational skills to overcome local minimum situations and navigate through complex scenarios. This highlights SQN's significant adaptability and strategic capability in end-to-end mapless navigation.
Notably, we observe that our SQN can leverage its adaptive decision mechanism for zero-shot transfer to previously unseen novel environments. These environments encompass scenarios with noisy disturbance conditions, observation noise settings, or unstructured observation patterns, such as open-space non-convex obstacles or subterranean cave-like scenes, demonstrating our method's capacity to handle out-of-distribution situations.

The summary of our contributions is as follows:
\begin{itemize}
    \item
    We present Skill Q-Network (SQN), a novel RL method capable of learning multiple navigation skills through adaptive skill ensemble.
    \item
    We design a tailored reward function to learn effective mapless navigation in complex environments.
    \item
    We empirically demonstrate the effectiveness of our adaptive skill ensemble method in addressing challenging mapless navigation problems across diverse environments including out-of-distribution settings.
\end{itemize}

%% file: sections/2.relatedworks.tex
\section{Related Works}
\label{sec:relatedworks}
\subsection{Conventional Mapless Navigation}
Various approaches have been developed for robotic navigation in unknown environments. Conventional methodologies~\cite{witting2018history,dang2019graph,lindqvist2021exploration} often employ Simultaneous Localization and Mapping (SLAM) to generate maps with topological graphs for navigating new terrains. These methods incrementally expand the map as the agent traverses the area, effectively managing the explored regions~\cite{yang2021graph,chen2022fast,kim2023topological}. This dynamic process enables the robot to distinguish between explored and unexplored regions in environments, thereby facilitating stable navigation.
However, map generation and graph management are both complex and computationally demanding processes. The complexity of these operations escalates significantly with the size of the environment, making it challenging to apply these solutions in expansive or complex areas. Moreover, they depend on handcrafted rules and the heuristics of human engineers to balance exploration in unknown spaces and goal-oriented navigation toward a destination. This reliance can hinder the agent's ability to generalize to new scenarios.

\subsection{Learning-based Mapless Navigation}
To overcome existing limitations, recent studies have focused on developing learning-based navigation policies that eliminate the need for map generation, utilizing imitation~\cite{pfeiffer2017perception,tsai2021mapless,yan2022mapless} and reinforcement learning~\cite{zhelo2018curiosity,fan2018crowdmove,grando2021deep,lv2021deep,dobrevski2021deep,sun2023risk,wang2023curriculum,wei2024memory,wu2020deep,zhao2023learning,huang2023creating,lee2023adaptive} approaches. Many of these studies have introduced mapless policy networks that process ego-centric sensory inputs, such as range measurements, and relative distance and orientation towards the goal. Initial research efforts often employed network architectures based on Multilayer Perceptrons (MLP), which solely rely on current observations for navigation~\cite{zhelo2018curiosity,fan2018crowdmove,grando2021deep,lv2021deep}. Although these models are suitable for simple scenarios, they often struggle to navigate complex environments with intricate topologies and fail to avoid getting stuck in unstructured regions. 

Some researchers have adopted recurrent neural network modules, such as Long Short-Term Memory (LSTM) or Gated Recurrent Units (GRU), for memory-based, mapless navigation~\cite{dobrevski2021deep,sun2023risk,wang2023curriculum,wei2024memory}. These methods utilize the temporal features of egocentric observations, facilitating efficient navigation in complex environments without the need to revisit areas. However, they primarily rely on a single policy network, which complicates the learning of adaptable maneuvers across various navigation scenarios, thereby limiting their ability to generalize in new environments.

Recent studies have introduced skill-based learning techniques~\cite{wu2020deep,zhao2023learning,huang2023creating,lee2023adaptive} for acquiring diverse navigation strategies through hierarchical learning. However, they still require separate high-level and low-level learning processes, necessitating tailored schemes for training each low-level skill~\cite{huang2023creating,lee2023adaptive}. Moreover, the utilization of predefined skills introduces heuristics that may compromise generalization performance in new situations. In contrast, our approach learns low-level policies and an adaptive skill combination mechanism without relying on skill-level priors, offering a novel pathway to address the limitations of existing models.

%% file: sections/4.methodologies.tex
\section{Methodologies}
\label{sec:methodologies}

\subsection{State and Action Representation}
To represent the surrounding navigation scene, our network receives two types of input state data, range measurements and navigational information.
The range measurements $o^{range}_t \!\in \mathbb{R}^{71}$ constitute a set of distances to occupied areas, each calculated from the ego robot's center point. The maximum observation range is 5 m, and the field-of-view angle spans $350\degree$ from left to right, with an angle resolution of $5\degree$.
The navigational information $o^{goal}_t \!\in \mathbb{R}^2$ contains the line-of-sight distance $d^{goal}_t$ and heading angle $\psi^{goal}_t$ toward the goal point. 
Both data types are ego-centric and do not require prior knowledge of the driving environment while navigating to the goal point. To handle large distances and accommodate environments of various sizes, we normalize all observation values to a range of 0 to 1.

Our agent is assumed to be a differential-wheeled robot controlled by twist commands $c = [v, \omega]$, which include translational ($v$ in m/s) and rotational ($\omega$ in rad/s) velocities. Accordingly, the robot agent's action space is represented by the following five discrete actions $a_t \in \mathbb{R}^5$: \textit{no-operation} ($[0, 0]$), \textit{forward} ($[2.5, 0]$), \textit{backward} ($[-1.25, 0]$), \textit{turn-left} ($[0, +\pi]$), and \textit{turn-right} ($[0, -\pi]$).

\subsection{Skill Q-Network}
To design a policy network capable of utilizing multiple skill policies, we incorporate the concept of functional modularity, as proposed in~\cite{seong2024self}, and develop the Skill Q-Network (SQN). The SQN is composed of modules categorized into three functions: the perception module, the planning module, and multiple latent skill policy modules (Fig. \ref{fig:network}).

The perception module extracts sensory features from two types of observations: $o^{range}_t$ and $o^{goal}_t$. These observations are individually transformed into $z^{range}_t \!\in \mathbb{R}^{128}$ and $z^{goal}_t \!\in \mathbb{R}^{128}$, respectively, using a Multi-layer Perceptron (MLP) that consists of two linear layers and ReLU non-linearity (Eq. \ref{eq:mlp_range_goal}). The resulting features are then concatenated into a single hidden vector. Additionally, to capture temporal dependencies on past observations, a GRU layer is integrated, enabling the network to derive a comprehensive perceptual feature $z^p_t \!\in \mathbb{R}^{128}$ that incorporates hidden features from previous observations $h_{t-1}$ (Eq. \ref{eq:gru}).
\begin{flalign}
    \label{eq:mlp_range_goal}
    z^i_t &= \text{MLP}_i(o^i_t), \quad i = \{range, goal\} \\
    \label{eq:gru}
    z^p_t, h_t &= \text{GRU}([ z^{range}_t; z^{goal}_t], h_{t-1})
\end{flalign}
\begin{figure}[t]
\centering
\includegraphics[width=0.42\textwidth]{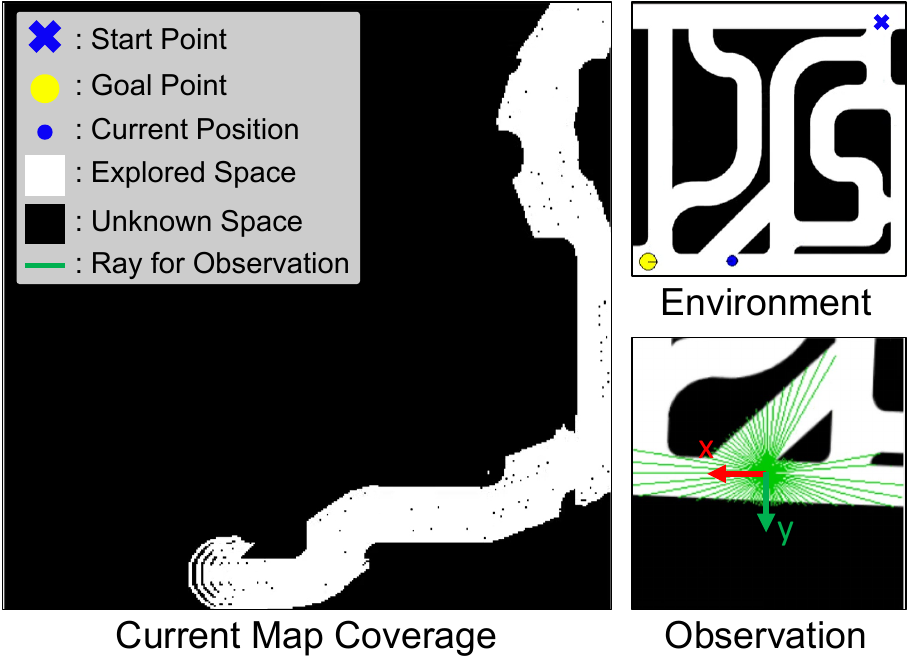}
\caption{
An example illustrating the map coverage, along with the corresponding status and observation of the ego agent.
}
\label{fig:map_cover}
\vspace{-2em}
\end{figure}

The decision module derives a skill decision as an attention score vector for multiple latent skill policies based on the hidden feature $z^p_t$. To promote functional modularity within the skill decision-making mechanism, we add a learnable module embedding $e^{dec}_{mod} \!\in \mathbb{R}^{128}$ to the input perceptual feature. This enhanced input is then processed by an MLP block, equipped with two linear layers and ReLU nonlinear activation (Eq. \ref{eq:mlp_dec}). Finally, we apply a softmax operation to generate the skill decision $\eta^{\pi}_{t} \!\in \mathbb{R}^{N}$, which indicates the importance of each latent skill policy output (Eq. \ref{eq:skill_dec}).
\begin{flalign}
    \label{eq:mlp_dec}
    z^{dec}_t &= \text{MLP}_{dec}(z^p_t + e^{dec}_{mod}), \quad e^{dec}_{mod} \in R^{128} \\
    \label{eq:skill_dec}
    \eta^{\pi}_{t} &= [\eta^{\pi_{1}}_t, ..., \eta^{\pi_{N}}_t] = \frac{\text{exp}(z^{dec}_t)}{\sum\nolimits_{i=1}^{N} \text{exp}(z^{dec}_t(i))}
\end{flalign}

The latent skill policy modules consist of N skill policy networks, each generating individual Q-values. Similar to the decision module, we apply module embeddings to distinguish the functional roles of the policies. Each latent skill policy is implemented using an MLP with two linear layers and ReLU non-linearity (Eq. \ref{eq:q_skill}). The Q-values produced by each skill network ($Q^{\pi_i}_t, \ i=1, …, N$) are aggregated into a single Q-value $Q^{\pi}_t$ through multiplication with the skill decision $\eta^{\pi}_t$ from the decision module (Eq. \ref{eq:q_final}). Finally, an epsilon-greedy strategy is employed to select the final discrete action $a_t \in \mathbb{R}^{m}$.
\begin{flalign}
    \label{eq:q_skill}
    Q^{\pi_{i}}_t \!&=\! \text{MLP}_{\pi_{i}}(z^p_t + e_{{mod}_{i}}), \quad i = 1, 2, …, N \\
    \label{eq:q_final}
    Q^{\pi}_t \!&=\! [\eta^{\pi_{1}}_t, ..., \eta^{\pi_{N}}_t]\! \times \!\! \begin{bmatrix}
    Q^{\pi_{1}}_t(s, a_1), \!\!\!\!& ..., \!\!\!\!& Q^{\pi_{1}}_t(s, a_m)\\
    & ... \\
    Q^{\pi_{N}}_t(s, a_1), \!\!\!\!& ..., \!\!\!\!& Q^{\pi_{N}}_t(s, a_m)\\
    \end{bmatrix}
\end{flalign}

Since SQN integrates a recurrent neural network, we adopt the R2D2\cite{kapturowski2018recurrent} reinforcement learning framework, which utilizes a burn-in mechanism. This approach involves executing inference for the initial steps of sequential episode data acquired through batch sampling. This process initializes the GRU's hidden state with the burn-in technique, enabling the computation of objectives based on Q-values under warm-start conditions. R2D2's training method effectively manages variable sequence states, significantly enhancing the stability and efficiency of learning from long-term sequential data. For further details on the training algorithm, we refer to \cite{kapturowski2018recurrent}.

\subsection{Reward Function}
\begin{figure}[t]
\centering
\includegraphics[width=0.45\textwidth]{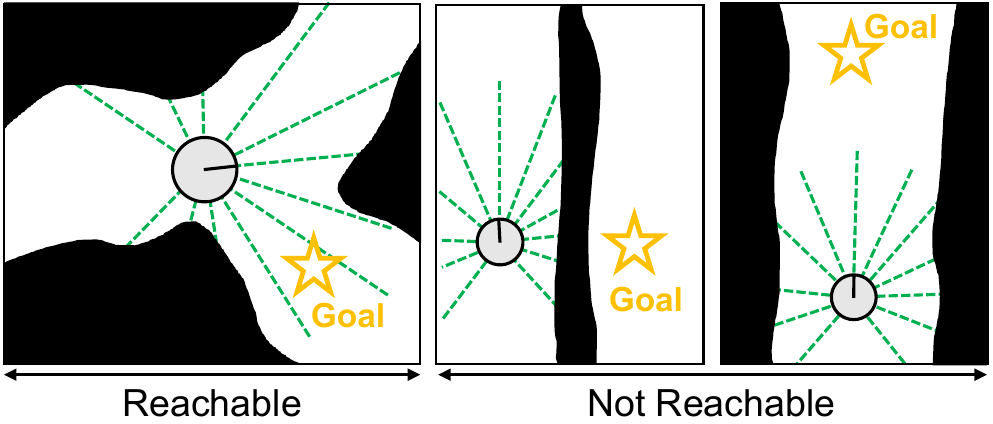}
\caption{
Examples of goal (non-)reachability. In mapless environments, situations frequently arise where the goal point is close in terms of Euclidean distance yet remains unreachable.
}
\label{fig:reachability}
\vspace{-1.5em}
\end{figure}
We designed two positive reward terms and one negative penalty term to train diverse navigation skills.

\textbf{Exploration Term:} We compute map coverage using range measurements to represent the explored area on the global map. Subsequently, we calculate the difference between the sizes of the current and subsequent explored areas to generate a positive reward signal, as follows:
\begin{flalign}
    \label{eq:reward_exp}
    r_{exp} = M(o^{range}_{t+1:0}) - M(o^{range}_{t:0}),
\end{flalign}
where $M(o^{range}_{t:0})$ represents the size of the map coverage up to the current observation $o^{range}_t$.
This term encourages the ego agent to explore new regions and enhances navigation progress toward the goal in unknown environments. Additionally, it discourages the robot from revisiting previously explored areas, thus promoting more efficient navigation. Note that the map coverage is used solely as privileged information within the reward signal and does not form part of the agent's input state. Consequently, it allows our model to harness the benefits of map generation-based approaches~\cite{kim2023topological} without the need to construct an expensive SLAM map.

\textbf{Reachability-Aware Navigation Term:}
We introduce a positive reward signal $r_{nav}$ for the agent's goal navigation. Typically, this signal is designed based on the Euclidean distance between the goal point and the ego robot~\cite{marchesini2020discrete,lv2021deep}. However, this method may excessively focus on minimizing the distance to the goal, leading to maneuvers that are susceptible to local minima. On the other hand, offering a sparse reward only~\cite{grando2021deep,wang2023curriculum} upon reaching the goal may encourage the acquisition of various maneuvers, but it tends to make learning navigation more complex compared to the use of dense rewards. Given these considerations, we propose a partially dense reward term that accounts for the distance to the goal point $g$, incorporating the reachability from the ego robot's current observation space $S_{obs}$ as follows:
\begin{flalign}
    \label{eq:reward_nav}
    r_{nav} =
    \begin{cases}
    \hfil exp(-d^{goal}_t) & {g} \in S_{obs} \\
    \hfil 0 & {g} \notin S_{obs}
    \end{cases},
\end{flalign}
where the goal $g$ is considered reachable (${g} \in S_{obs}$) if it is within the agent's field-of-view and the distance $d^{goal}_t$ is less than the observed range measurement in the direction of the goal.
This approach prevents the generation of over-optimistic reward signals in situations where the Euclidean distance to the goal point $g$ is small, but the goal is located beyond a non-traversable area (see Fig. \ref{fig:reachability}). As a result, it enables the ego agent to concentrate on learning exploration maneuvers by the signal $r_{exp}$, which facilitates escaping from local minimum situations through local exploration. Conversely, when the goal point is accessible to the ego robot, this reward term produces a positive signal, encouraging the agent to learn goal-directed maneuvers that prioritize reaching the goal over exploring new environments.

\textbf{Time Step Penalty Term:} Lastly, we define a constant negative reward term, $r_{time} = -1$, to incentivize the agent to perform actions that yield more positive reward signals and compensate for this negative penalty.

The total reward function is a weighted summation of the above reward terms as follows:
\begin{flalign}
    \label{eq:reward_total}
    r_t = 2.0 \times r_{nav} + 1.0 \times r_{exp} + 0.2 \times r_{time}
\end{flalign}

%% file: sections/5.experiment.tex
\section{Experiments}
\label{sec:experiment}
\begin{figure}[t]
\centering
\includegraphics[width=0.48\textwidth]{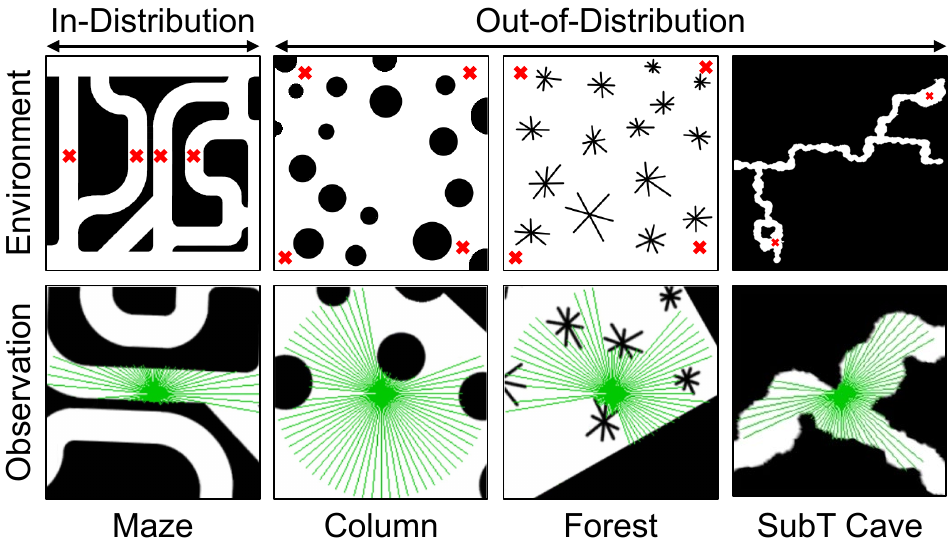}
\caption{
Various environments and observation patterns for experiments. Observations are visualized with the ego agent's center point as the origin, where the local x-axis points east.
}
\label{fig:environment}
\vspace{-1.5em}
\end{figure}

\begin{table*}[t] 
\caption{Performance Results in a Known Scenario (Maze) and Three Unseen Environments (Column, Forest, SubT Cave)}
\makegapedcells
\centering
\begin{adjustbox}{width=0.98\textwidth}
\Large
\renewcommand{\arraystretch}{0.3}
\begin{tabular}{c c c | c c c c c c}
\toprule
&\multicolumn{2}{c|}{{Maze}} 
&\multicolumn{2}{c}{{Column}}
&\multicolumn{2}{c}{{Forest}}
&\multicolumn{2}{c}{{SubT Cave}} \\

\cmidrule(lr){2-3}
\cmidrule(lr){4-5}
\cmidrule(lr){6-7}
\cmidrule(lr){8-9}
Method
    & {\makecell{Success (\%) $\uparrow$}}
        & {\makecell{Time Steps (-) $\downarrow$}}
    & {\makecell{Success (\%) $\uparrow$}}
        & {\makecell{Time Steps (-) $\downarrow$}}
    & {\makecell{Success (\%) $\uparrow$}}
        & {\makecell{Time Steps (-) $\downarrow$}}
    & {\makecell{Success (\%) $\uparrow$}}
        & {\makecell{Time Steps (-) $\downarrow$}} \\
\midrule
DQN   &
\text{0.72} $\pm$ \text{0.04} & \text{252.98} $\pm$ \text{10.46} &
\text{0.95} $\pm$ \text{0.03} & \text{151.49} $\pm$ \text{12.30} &
\textbf{0.97} $\pm$ \textbf{0.01} & \textbf{167.00} $\pm$ \textbf{\; 6.74} &
\text{0.38} $\pm$ \text{0.03} & \text{1061.35} $\pm$ \text{16.04} \\
R2D2   &
\text{0.70} $\pm$ \text{0.04} & \text{257.82} $\pm$ \text{10.85} &
\textbf{0.99} $\pm$ \textbf{0.01} & \text{172.44} $\pm$ \text{\; 1.59} &
\text{0.91} $\pm$ \text{0.03} & \text{221.78} $\pm$ \text{10.78} &
\text{0.64} $\pm$ \text{0.02} & \text{\; 937.68} $\pm$ \text{11.76} \\
SQN   &
\textbf{0.98} $\pm$ \textbf{0.01} & \textbf{181.10} $\pm$ \textbf{\; 5.18} &
\textbf{0.99} $\pm$ \textbf{0.01} & \textbf{148.67} $\pm$ \textbf{\; 2.24} &
\text{0.93} $\pm$ \text{0.02} & \text{171.43} $\pm$ \text{\; 6.17} &
\textbf{0.82} $\pm$ \textbf{0.03} & \textbf{\; 767.51} $\pm$ \textbf{32.30} \\
\bottomrule

\end{tabular}
\end{adjustbox}
    \label{table:eval}
\end{table*}

\begin{figure*}[t]
\centering
\includegraphics[width=0.98\textwidth]{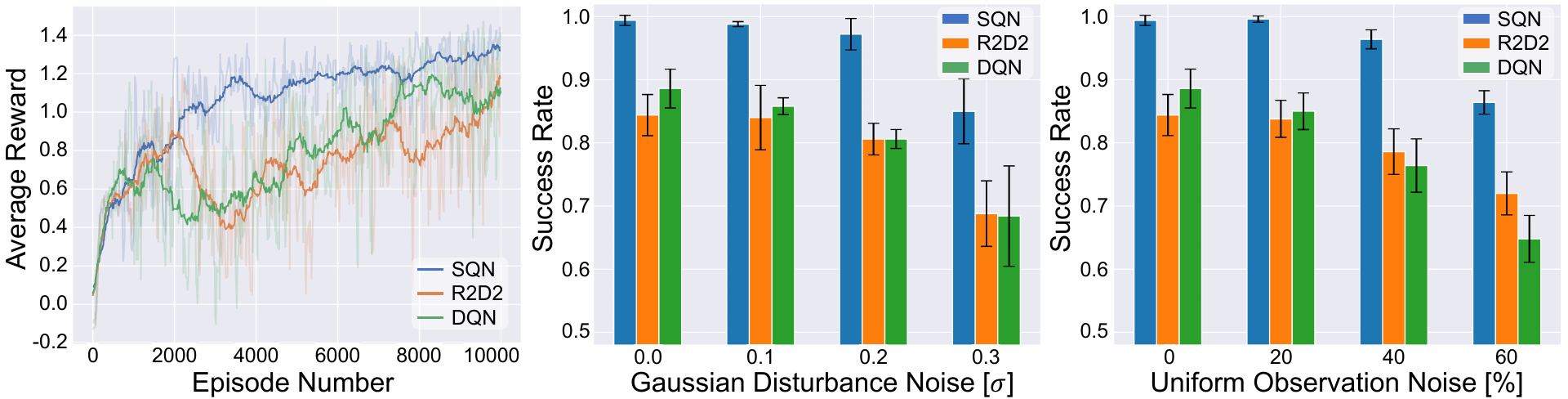}
\caption{
\textbf{Left: } Learning curves of the average reward in evaluation.
\textbf{Center: } Performance of different evaluation policies with external disturbance noise applied to the agent's motion.
\textbf{Right: } Performance of the policies with uniform noise added to the agent's observations.
}
\label{fig:result_perform}
\vspace{-1.0em}
\end{figure*}

\subsection{Environments Setup}
We utilize the Multiagent Particle Environments (MPE)~\cite{mordatch2018emergence} in the Pettingzoo framework~\cite{terry2021pettingzoo} to create a particle agent-based training environment. Differing from the original setup, we limit the robot's movements to forwards, backwards, and rotations, mirroring the physical constraints of differential-wheeled robots. In the simulation, we only update to the next state if it is traversable; otherwise, we stop the motion to avoid navigating into occupied areas of the global map. Each training episode is composed of 400 time steps, with each step lasting 0.1 seconds, based on a 10 Hz cycle for the ego robot's control system.

We train the policy network in the Maze environment and validate it against challenging scenarios within this environment. Additionally, we create three unseen environments (Column, Forest, SubT Cave) to assess the policy model's performance under OOD conditions.
During evaluation, start and goal points are randomly chosen from either the four or two highlighted locations (marked by red 'x's) in Fig. \ref{fig:environment}, focusing on scenarios that challenge the ego robot to skillfully navigate to the goal point.

\textbf{Maze: }
For training and evaluation, we construct the Maze environment ($17 \times 17$ m), featuring complex topology and multiple intersections. Various spawn points are defined for training. We randomly select two non-overlapping points to establish the ego robot's initial pose and goal point, with both position and orientation being randomly initialized. During evaluation, however, we focus on the four highlighted locations that can yield challenging routes, including multiple intersections between the start and goal points.

\textbf{Column: }The Column scenario is different from the training environment, being an open-space scenario with round-shaped obstacles in an unseen environment ($17 \times 17$ m). The ego robot primarily receives range observations with a convex pattern. There is a wide open space in the center that was not observed in the training environment.

\textbf{Forest: }The Forest is also an open-space environment ($17 \times 17$ m) and a new scenario with irregular asterisk-shaped obstacles. The ego robot primarily observes range measurements with a non-convex pattern. The asterisk-shaped terrain makes it easy for the ego agent to become trapped.

\textbf{SubT Cave: }This environment is an out-of-distribution, subterranean (SubT), large-scale ($51 \times 51$ m), cave-like setting with numerous non-convex terrains~\cite{kim2023topological}. The environment contains several forks that often lead the agent into local minimum situations or dead ends. With its realistic terrain, taking a wrong turn into a branch can significantly delay navigation toward the desired goal point. Considering that the spatial scale is three times larger than that of the other environments, we set the maximum number of time steps for each episode in this environment at 1200.

\subsection{Robustness Settings}
To evaluate the potential of our trained model for broader navigation tasks, such as navigating through noisy real-world environments, we assess the robustness of our agent by testing its ability to generalize across various modified environments. We deploy the model, trained in the Maze environment, to the following settings without additional retraining: 1) a setting where external Gaussian disturbance is randomly applied to the robot agent's translational and rotational velocities, and 2) a setting where uniform noise is applied to the agent's range measurement observations.

\subsection{Hyperparameter Configuration}
We train the SQN model with the following hyperparameters: the batch size is 64, and the layer dimensions in the SQN are all 128.
The target network is updated by Polyak updates with a coefficient $\tau=0.002$. The initial epsilon value is $\epsilon = 0.4$ and linearly decreases to the terminal value of $\epsilon = 0.1$. When updating the network, we sample the batch of 400-step sequence data and perform the \textit{burn-in} of 10 steps for each sequence. We train the network for a maximum of 10,000 episodes.
Considering the complexity of the environment and the number of expected strategies, we set the number of latent skill policy modules to 2 ($N=2$).

\subsection{Baseline Models}
\begin{itemize}
    \item
    \textbf{DQN: }Consistent with the methods in various studies~\cite{zhelo2018curiosity,marchesini2020discrete,grando2021deep}, this policy relies solely on current state inputs and employs the same discrete actions as SQN. To compensate for the network capacity difference compared to SQN, which consists of one decision module and two skill policy modules, we increased the size of the DQN network layer after the feature extraction module to three times that of SQN.
    \item 
    \textbf{R2D2: }Similar to previous studies~\cite{dobrevski2021deep,sun2023risk,wei2024memory}, this method incorporates a recurrent layer to capture temporal features. It shares the same backbone network structure as SQN's GRU-based perception module, followed by a single Q-value head. R2D2 is trained following the methodology outlined in \cite{kapturowski2018recurrent}, using the same sequence length and burn-in steps as SQN.
    
\end{itemize}

\section{Quantitative Evaluation}
\label{sec:quantitative}
\begin{figure*}[t]
\centering
\includegraphics[width=0.96\textwidth]{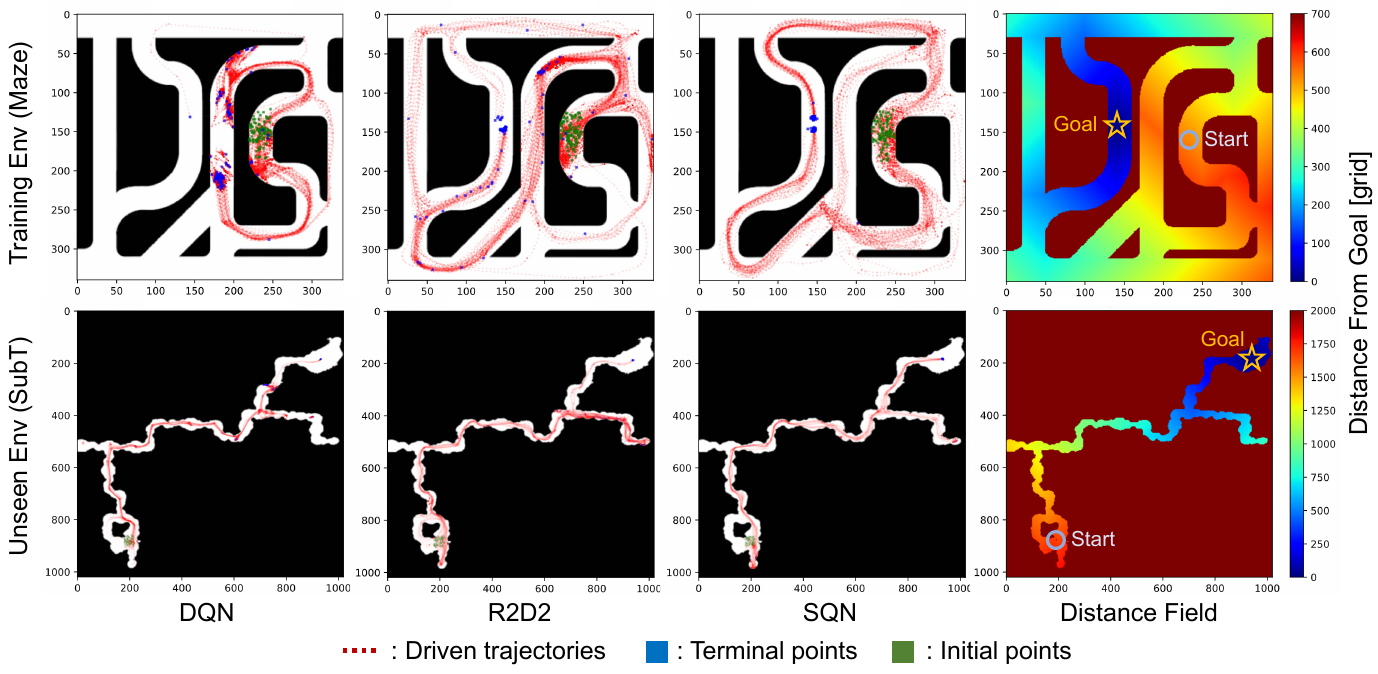}
\caption{
Trajectory results of evaluation policies in the evaluation environments, with distance fields visualizing the true distance cost from the goal point.
}
\label{fig:result_traj}
\vspace{-1.5em}
\end{figure*}

\subsection{Learning Curves}
Fig. \ref{fig:result_perform} (Left) illustrates the learning curves of the average reward for each policy model. The results indicate that SQN outperforms R2D2 and DQN by $14\%$ and $23\%$, respectively, achieving a final average reward of 1.35. R2D2 initially surpasses DQN, maintaining a higher performance until the first 2000 episode before experiencing a temporary decline. It recovers to a performance level of 1.18 after episode 3100. Similarly, DQN shows improvement until episode 1500, after which it faces a temporary drop, ultimately reaching a performance level of 1.10. The temporary performance drops observed in both models can be attributed to the evaluation scenarios, which feature routes prone to leading into local minimum regions before reaching the goal point. These drops can result from the models' predominant tendency to learn a single policy, causing temporary overfitting in areas characterized by local minima. Unlike these methods, our SQN model consistently demonstrates performance improvements without significant declines, benefiting from the adaptive ensemble of skill policies.

\subsection{Performance Comparison}
For the quantitative comparison, we evaluate two metrics: Success (success rate) and Time Steps (the number of steps taken in an episode). We assess the success rate of reaching the goal point for all policies across 100 episodes with 5 different seeds, as well as the number of steps taken to reach the goal. We then calculate the mean and standard deviation of these metrics.
In the Column, Forest, and SubT Cave environments, characterized by OOD observations and scenarios distinct from Maze, we deploy evaluation policies via zero-shot transfer, without the need for additional training.

Table \ref{table:eval} summarizes the overall performance of the policy models.
In challenging scenarios within the Maze environment, our SQN demonstrates superior navigation performance, achieving a success rate up to 40\% higher (0.98) and requiring up to 42\% fewer time steps (181.20 steps) compared to baseline models. Despite being trained in the Maze, the models still encounter challenges, mainly due to the numerous intersections between start and goal points. These challenges result in lower performance for the two baseline models equipped with a single policy network pipeline, with success rates around 0.70 and requiring over 250 steps to complete.
In Column and Forest environments, where there are no dead ends between the start and goal, all the policies exhibit high success rates exceeding 0.90.
In SubT Cave, characterized by uneven terrain requiring long-term exploration, the DQN, relying solely on current states, exhibits the lowest performance at 0.38. R2D2, utilizing sequential capabilities alongside exploration and reachability-based reward signals, achieves performance exceeding 0.50 in the novel environment but is capped at 0.64. In contrast, our SQN achieves remarkable zero-shot transfer performance of 0.82 even when deployed in scenarios with OOD range measurements, owing to its skill-based navigational policy.
\begin{figure*}[t]
\centering
\includegraphics[width=0.99\textwidth]{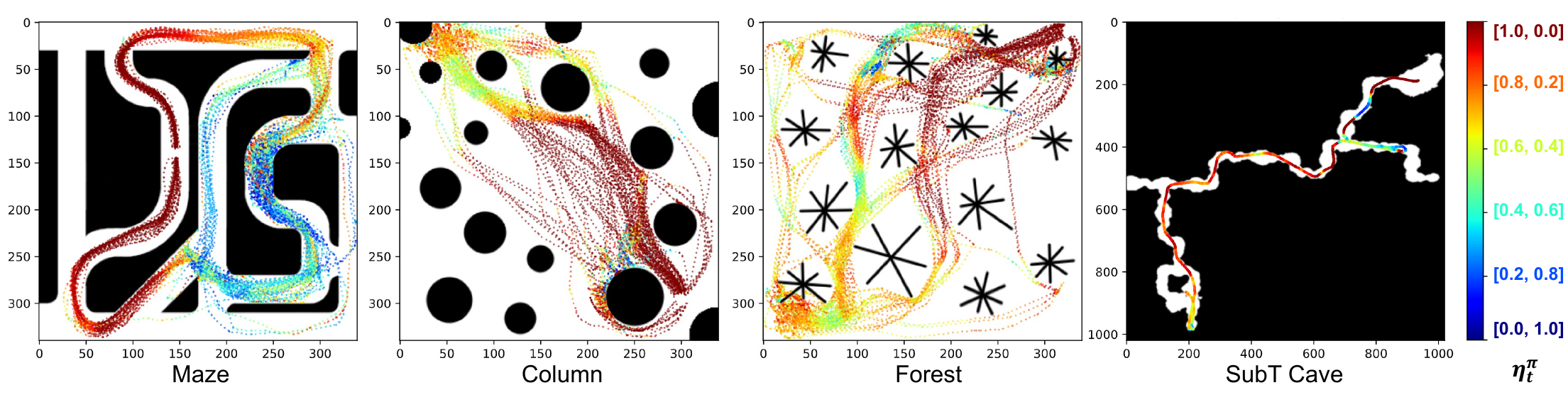}
\caption{
Skill trajectories of SQN across various environments. For brevity, a single trajectory is represented in SubT Cave.
}
\label{fig:result_dec}
\vspace{-1.5em}
\end{figure*}

\subsection{Robustness}
We analyze the robustness of policy models in the Maze environment under two sets of modified conditions. All experiments are conducted with 5 seeds, with each run consisting of 100 episodes to assess the success rate.

\subsubsection{External Disturbance}
Fig. \ref{fig:result_perform} (Center) shows the performance variation of the evaluation policies when Gaussian disturbance noise with a mean of 0 is applied to the ego agent's translational and rotational velocity. We evaluate the policies across three settings ($\sigma\!=0.1, 0.2, 0.3$) by incrementing the noise's standard deviation, $\sigma$, by 0.1. The results for the $\sigma\!=0.0$ setting represent disturbance-free baseline performance. Our SQN closely matches the disturbance-free result up to the $\sigma\!=0.2$ setting, demonstrating the highest success rate of 0.97. When subjected to a strong disturbance with $\sigma\!=0.3$, our SQN maintains a high performance level of up to 0.85. In contrast, the two baseline models experience a significant drop in performance, falling below 0.70.

\subsubsection{Observation Noise}
Fig. \ref{fig:result_perform} (Right) depicts the performance changes when different degrees of uniform noise are applied to the ego agent's range measurements. We increase the magnitude of the uniform noise by $20\%$ and evaluate the policies at three different settings ($20\%, 40\%, 60\%$), compared to the noise-free performance ($0\%$).
The result of SQN remains close to that of the noise-free setting up to a noise level of $40\%$, achieving the highest success rate of 0.96. In contrast, the other two baseline models exhibit a significant drop in performance as the magnitude of observation noise increases, marked by a large standard deviation in their performance results. SQN sustains a performance level exceeding 0.80 with a minimal deviation of 0.02, even at the highest noise level of $60\%$.

\section{Qualitative Analysis}
\label{sec:qualitative}
\subsection{Learned Navigation Behavior}
Fig. \ref{fig:result_traj} visualizes the navigation trajectories of the policies in the environments, Maze and SubT Cave, accompanied by the corresponding distance fields. These fields depict the actual distance to the goal point in each evaluation scenario.

Referring to the distance field, we observe that SQN primarily navigates towards regions that reduce the distance cost to the goal. This result demonstrates that SQN achieves near-optimal navigation performance in the complex Maze environment.
In the case of DQN, the method lacks access to previously encountered features, leading it to navigate solely towards the goal point. This often results in the DQN becoming trapped in areas of local minima without reaching the goal.
R2D2 can access the previous state features during navigation, achieving a higher success rate than DQN. However, with only a single Q-head, the model struggles to learn a policy that can handle the diverse intersections encountered en route to the goal. This often results in unnecessary revisits and failure to reach the goal point within the maximum allowable steps.
Leveraging the skill decision, SQN can employ various skill policies to efficiently navigate complex topological regions en route to the goal point. It also demonstrates the ability to quickly detour and reach the goal, even after entering sub-optimal areas.

In the SubT Cave environment, which features unseen terrain observations, the performance gap between SQN with its adaptive skill ensemble mechanism and other baseline models is more pronounced.
On this map, a fork near the (700, 400) location leads to two branches before the upper-right goal point. Choosing the wrong branch leads to a dead end, requiring a long-term return maneuver not encountered in the training scenarios. In such an environment, DQN suffers from the unstructured rough terrain near the fork, often leading to it getting stuck and failing to reach the goal point. R2D2, leveraging the ego robot's state sequence, is more proactive than DQN. However, its frequent oscillation between the fork and the dead end, along with occasional failures to exit the dead end, results in few successful trajectories to the goal. SQN, on the other hand, demonstrates efficient navigation with minimal back-and-forth maneuvers before arriving at the goal. It also performs swift return maneuvers to the fork without unnecessary revisits, even when encountering a dead end, thereby navigating towards the correct branch leading to the goal point.

\subsection{Learned Skill Trajectories}
To analyze how SQN utilizes its learned skills, we visualize the trajectories of the skill decision while navigating the four evaluation environments with fixed goal points (Fig. \ref{fig:result_dec}). The results show that our SQN can learn diverse combinations of latent low-level skills through adaptive skill ensemble.
In the Maze environment, SQN initially produces skill decisions focusing more on the second latent skill policy, $\pi_2$. These decisions enable the agent to perform exploration maneuvers to escape from the initial spawn point (near the (235, 162) location), even as the Euclidean distance to the goal point increases. Upon reaching regions near the bottom-left or top-right corner, SQN shifts its emphasis to skill decisions that prioritize the first latent skill policy, $\pi_1$, facilitating goal-directed maneuvers.
In the Column and Forest maps, when the agent is randomly initialized at the start point (Column: top-left, Forest: bottom-left), SQN generates decisions that balance the two skill policies. Once the route toward the goal becomes more straightforward, our method transitions to prioritizing the first skill, $\pi_1$.
Furthermore, upon encountering obstacle areas near the goal point, it demonstrates adaptability by temporarily making decisions that highlight more on $\pi_2$ to navigate out of the blocked areas.
In the SubT Cave episode, the ego agent infers decisions that give more weight to $\pi_1$ when navigating towards the goal point along uneven terrain. However, the agent adjusts its decisions to increase the weight on $\pi_2$ in situations where it needs to return to the fork after taking the branch leading to a dead end.
These results demonstrate that our SQN can learn an adaptive skill decision mechanism that ensembles latent low-level policies, effectively handling multiple local minima and sub-optimal situations encountered while navigating new environments.

%% file: sections/6.conclusion.tex
\section{Conclusion}
\label{sec:conclusion}
In this paper, we introduce Skill Q-Network, a policy network method based on skill ensemble mechanisms. We present a tailored reward function designed to learn exploration and goal-directed navigation strategies in mapless environments.
Our proposed method demonstrates remarkable navigation performance compared with those of other baseline models in four complex unstructured scenarios. 
Furthermore, our empirical experiments demonstrate the adaptability and robustness of our method when transferred to novel OOD environments in a zero-shot manner. These environments include scenarios such as open spaces with non-convex obstacles or uneven terrain with multiple branches and dead ends.
In future work, we aim to explore the capabilities of our skill ensemble mechanism to achieve not only observation-level generality but also dynamics feature-level versatility, leveraging domain randomization schemes~\cite{peng2018sim,mehta2020active}. This extension will allow us to further investigate and address the challenge of robust sim-to-real transfer.